\documentclass[conference]{IEEEtran}

\makeatletter
\def\ps@IEEEtitlepagestyle{%
  \def\@oddfoot{\mycopyrightnotice}%
  \def\@evenfoot{}%
}
\def\mycopyrightnotice{%
  {\footnotesize 978-1-6654-7095-7/22/\$31.00~\copyright~2022 IEEE\hfill}
  \gdef\mycopyrightnotice{}
}

\usepackage{blindtext}
\usepackage{eso-pic}
\IEEEoverridecommandlockouts
\usepackage{cite}
\usepackage{amsmath,amssymb,amsfonts}
\usepackage{algorithmic}
\usepackage{graphicx}
\usepackage{textcomp}
\usepackage{xcolor}

\usepackage{subfig}

\def\BibTeX{{\rm B\kern-.05em{\sc i\kern-.025em b}\kern-.08em
    T\kern-.1667em\lower.7ex\hbox{E}\kern-.125emX}}
    
\usepackage{eso-pic}
\newcommand\AtPageUpperMyright[1]{\AtPageUpperLeft{%
 \put(\LenToUnit{0.17\paperwidth},\LenToUnit{-2cm}){%
     \parbox{0.9\textwidth}{\raggedleft\fontsize{8}{11}\selectfont #1}}%
 }}%
\newcommand{\conf}[1]{%
\AddToShipoutPictureBG*{%
\AtPageUpperMyright{#1}
}
}    

\usepackage{booktabs}
\usepackage{listings}
\usepackage{scalerel}
\usepackage[binary-units]{siunitx}
\usepackage{tablefootnote}
\usepackage[textsize=tiny]{todonotes}
\usepackage{tikz, pgfplots}
  \usetikzlibrary{svg.path}

\makeatletter
\let\NAT@parse\undefined
\makeatother
\usepackage{hyperref}

\newcommand\httpsurl[1]{%
  \href{https://#1}{\nolinkurl{#1}}%
}

\definecolor{orcidlogocol}{HTML}{A6CE39}
\tikzset{
  orcidlogo/.pic={
    \fill[orcidlogocol] svg{M256,128c0,70.7-57.3,128-128,128C57.3,256,0,198.7,0,128C0,57.3,57.3,0,128,0C198.7,0,256,57.3,256,128z};
    \fill[white] svg{M86.3,186.2H70.9V79.1h15.4v48.4V186.2z}
                 svg{M108.9,79.1h41.6c39.6,0,57,28.3,57,53.6c0,27.5-21.5,53.6-56.8,53.6h-41.8V79.1z M124.3,172.4h24.5c34.9,0,42.9-26.5,42.9-39.7c0-21.5-13.7-39.7-43.7-39.7h-23.7V172.4z}
                 svg{M88.7,56.8c0,5.5-4.5,10.1-10.1,10.1c-5.6,0-10.1-4.6-10.1-10.1c0-5.6,4.5-10.1,10.1-10.1C84.2,46.7,88.7,51.3,88.7,56.8z};
  }
}
\newcommand\orcidicon[1]{\href{https://orcid.org/#1}{\mbox{\scalerel*{
\begin{tikzpicture}[yscale=-1,transform shape]
\pic{orcidlogo};
\end{tikzpicture}
}{|}}}}

\newcommand\copyrighttext{%
    \scriptsize \copyright{ }2022 IEEE. Personal use of this material is permitted. Permission from IEEE must be obtained for all other uses, in any current or future media, including reprinting/republishing this material for advertising or promotional purposes, creating new collective works, for resale or redistribution to servers or lists, or reuse of any copyrighted component of this work in other works.}
\newcommand\copyrightnotice{%
    \begin{tikzpicture}[remember picture,overlay]
    \node[anchor=south,yshift=10pt,xshift=7pt] at (current page.south) {\parbox{\dimexpr\textwidth-\fboxsep-\fboxrule\relax}{\copyrighttext}};
    \end{tikzpicture}%
}

\begin{document}
\bstctlcite{IEEEexample:BSTcontrol}

\title{\vspace*{1cm} Data-Driven Occupancy Grid Mapping using Synthetic and Real-World Data*\\
\thanks{*This research is accomplished within the project ”UNICARagil” (FKZ 16EMO0284K). We acknowledge the financial support for the project by the Federal Ministry of Education and Research of Germany (BMBF).}
}

\author{\IEEEauthorblockN{Raphael van Kempen\textsuperscript{\orcidicon{0000-0001-5017-7494}}}
\IEEEauthorblockA{\textit{Institute for Automotive Engineering} \\
\textit{RWTH Aachen University}\\
Aachen, Germany \\
raphael.vankempen@ika.rwth-aachen.de}
\and
\IEEEauthorblockN{Bastian Lampe\textsuperscript{\orcidicon{0000-0002-4414-6947}}}
\IEEEauthorblockA{\textit{Institute for Automotive Engineering} \\
\textit{RWTH Aachen University}\\
Aachen, Germany \\
bastian.lampe@ika.rwth-aachen.de}
\and
\IEEEauthorblockN{Lennart Reiher\textsuperscript{\orcidicon{0000-0002-7309-164X}}}
\IEEEauthorblockA{\textit{Institute for Automotive Engineering} \\
\textit{RWTH Aachen University}\\
Aachen, Germany \\
lennart.reiher@ika.rwth-aachen.de}
\and
\IEEEauthorblockN{Timo Woopen\textsuperscript{\orcidicon{0000-0002-7177-181X}}}
\IEEEauthorblockA{\textit{Institute for Automotive Engineering} \\
\textit{RWTH Aachen University}\\
Aachen, Germany \\
timo.woopen@ika.rwth-aachen.de}
\and
\IEEEauthorblockN{Till Beemelmanns\textsuperscript{\orcidicon{0000-0002-2129-4082}}}
\IEEEauthorblockA{\textit{Institute for Automotive Engineering} \\
\textit{RWTH Aachen University}\\
Aachen, Germany \\
till.beemelmanns@ika.rwth-aachen.de}
\and
\IEEEauthorblockN{Lutz Eckstein}
\IEEEauthorblockA{\textit{Institute for Automotive Engineering} \\
\textit{RWTH Aachen University}\\
Aachen, Germany \\
lutz.eckstein@ika.rwth-aachen.de}
}

\maketitle
\conf{\textit{  Proc. of the International Conference on Electrical, Computer, Communications and Mechatronics Engineering  (ICECCME) \\ 
16-18 November 2022, Maldives}}

\copyrightnotice

\begin{abstract}
In perception tasks of automated vehicles (AVs) data-driven have often outperformed conventional approaches. This motivated us to develop a data-driven methodology to compute occupancy grid maps (OGMs) from lidar measurements. Our approach extends previous work such that the estimated environment representation now contains an additional layer for cells occupied by dynamic objects. Earlier solutions could only distinguish between free and occupied cells. The information whether an obstacle could move plays an important role for planning the behavior of an AV. We present two approaches to generating training data. One approach extends our previous work on using synthetic training data so that OGMs with the three aforementioned cell states are generated. The other approach uses manual annotations from the nuScenes~\cite{Caesar.2020} dataset to create training data. We compare the performance of both models in a quantitative analysis on unseen data from the real-world dataset. Next, we analyze the ability of both approaches to cope with a domain shift, i.e. when presented with lidar measurements from a different sensor on a different vehicle. We propose using information gained from evaluation on real-world data to further close the reality gap and create better synthetic data that can be used to train occupancy grid mapping models for arbitrary sensor configurations. Code is available at \url{https://github.com/ika-rwth-aachen/DEviLOG}.
\end{abstract}

\begin{IEEEkeywords}
AD, perception, simulation, deep learning
\end{IEEEkeywords}


\section{INTRODUCTION}

An automated vehicle can determine the drivable space in the static environment by localizing itself on a high-definition map. These HD maps describe the exact road geometry and traffic rules applying to each lane. 

\begin{figure}
\centering
\subfloat[Trained on Synthetic Data]{
	   \includegraphics[width=0.45\columnwidth]{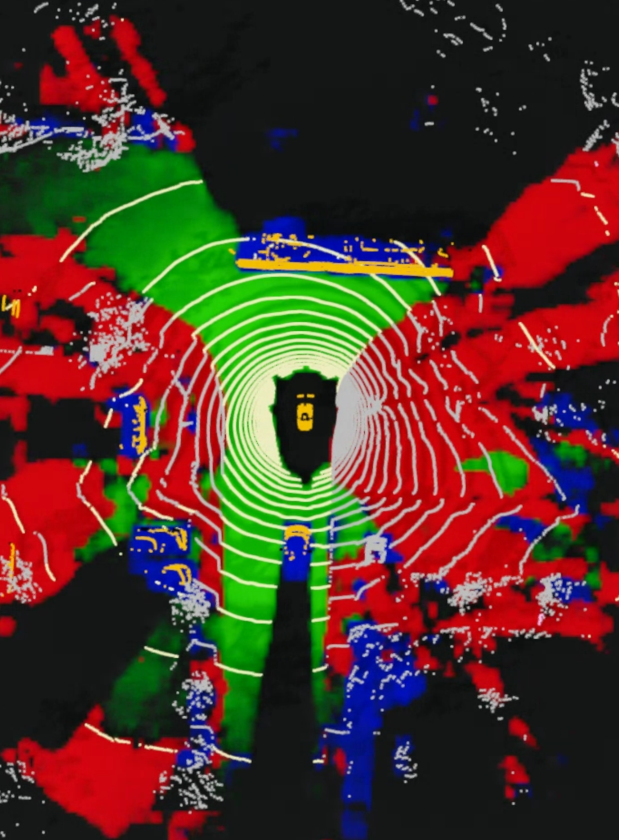}\label{fig:prediction-sim}
	}
\subfloat[Trained on nuScenes]{
	\includegraphics[width=0.45\columnwidth]{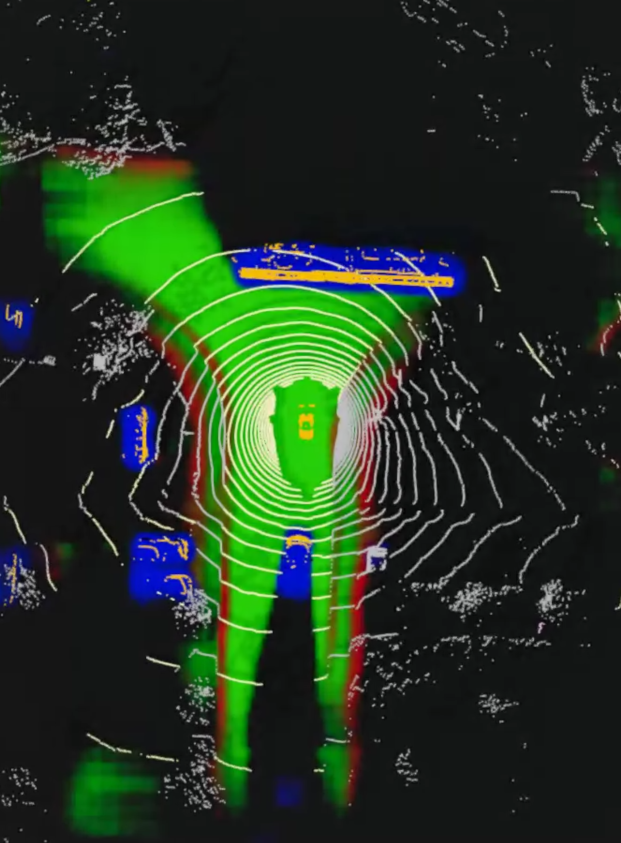}\label{fig:prediction-nuscenes}
	}
\caption{The left OGM was predicted by a model trained on synthetic data and the right OGM was predicted by a model trained on labels generated from annotations of the nuScenes \cite{Caesar.2020} dataset. Green indicates belief mass for \textit{free}, red for \textit{statically occupied} and blue for \textit{dynamically occupied} cells. The lidar point cloud as input data is superimposed and colored by the class annotations, i.e. vehicles are orange, drivable surface is yellow and occupied space is grey.}
\label{fig:predictions}
\end{figure}

However, these maps can be outdated so that it becomes necessary to substitute them with measurement-based representations to determine the actual drivable space. Grid-based environment representations are particularly suitable for this as they discretize a defined area around the vehicle into cells that can be aligned with the HD map. Distances measured e.g. by a lidar sensor can be used to assign an occupancy state to each cell in an OGM. Such OGMs are also used in the UNICARagil project~\cite{Buchholz.2020, Woopen.2018}, where our approach is also developed. The trajectories planned by an AV are constrained by static obstacles (e.g. road boundaries, barricades) as well as potentially dynamic obstacles (e.g. cars, pedestrians). While static obstacles always constrain the drivable space, the impact of dynamic obstacles depends on their future motion. Hence it is necessary to predict their future behavior. This is often achieved using object-based approaches estimating the motion states of objects in the vehicle's environment. Several approaches for detection, tracking and prediction of objects using cameras, lidar and radar sensors have been published in the past years. However, object-based algorithms are often trained on a fixed set of classes and objects that do not appear in the training dataset may not be detected or falsely classified. On the other hand, OGMs can represent obstacles of arbitrary shape and class. For object-based detectors a threshold usually needs to be found that minimizes both the false-positive-rate and the false-negative-rate. If free space can be estimated with high certainty using an OGM, which is combined with the object-based representation, it is possible to allow a higher false-positive-rate to further reduce the false-negative-rate and the robustness of perception can be increased. Semantic grid maps (SGMs) extend the task of occupancy grid mapping to assigning semantic classes (e.g. road, sidewalk, pedestrian, car) to each cell \cite{Bieder.2020, Richter.2020}. While the aforementioned works focus on creating semantic grid maps for a wide variety of classes, we extended our previous approach~\cite{vanKempen.2021b} to the three classes \textit{free}, \textit{statically occupied} and \textit{dynamically occupied}. Our goal is to estimate OGMs with very high performance on these classes and to increase the robustness of perception by fusing them with object-based representations. As detecting and estimating the actual motion of objects is a more complex task, we want to separate this into another module to increase the robustness of the module developed in this work. To enable sensor fusion and cooperative perception \cite{Lampe.2020}, it is necessary to quantify uncertainties in the outputs of a perception algorithm, which is implemented in our approach by predicting evidential OGMs.

A conventional way to generate OGMs is using a geometric inverse sensor model (ISM) with range measurements \cite{Thrun.2005}. A grid map obtained from one measurement is usually quite sparse and does not contain semantic information. A denser OGM including dynamic information can be obtained by tracking it over time \cite{Nuss.2018} or by using deep ISMs, which use supervised learning to estimate OGMs or SGMs from measurements. Some approaches use cross-modal training \cite{Bauer.2019b} but suffer from the restrictions of the underlying geometric ISM as the generated labels are just an estimation of the ground-truth. Other works use manually annotated datasets \cite{Bieder.2020, Lee.03.08.2020, Fei.2021b, Schreiber.2021b} to generate ground truth OGMs or SGMs, e.g. using semantically labeled point clouds or bounding boxes. These manually annotated datasets have to be created with high effort and can only represent the sensor setup and environment conditions that were recorded. Our approach presented in \cite{vanKempen.2021b} uses advanced sensor models to generate synthetic training data to train a deep neural network to predict OGMs from lidar measurements, which outperforms a geometric approach. This approach allows generating training data for arbitrary sensor configurations with low effort. While the presented qualitative analysis on real-world data was already quite promising, a quantitative analysis on real-world data was still due and is presented in this work among other things.

In this work, we extend our previous approach to predict OGMs with cell states \textit{free}, \textit{statically occupied} and \textit{dynamically occupied}. Additionally, we provide a quantitative performance analysis on the real-world nuScenes dataset~\cite{Caesar.2020}. The performance of the model trained on synthetic data is compared to a model trained on labels generated from the manual annotations of the dataset. Finally, we compare the transferability of both approaches to a new sensor configuration by applying them to one of our research vehicles. As a result, we propose a methodology of using real-world datasets to identify blind spots in the simulation and improve the domain adaptation of models trained on synthetic data to real-world measurements. By making our baseline and evaluation algorithm open-source, we enable others to compare their approaches with ours.

\section{BACKGROUND}

This section gives an overview of current approaches for computing OGMs and SGMs. Evidence theory, the basis for the output of the proposed neural network, is briefly introduced and we will give an short overview on the state of the art for using synthetic training data for automotive perception tasks.

\subsection{Occupancy Grid Maps}\label{ogm}

An OGM \cite{Elfes.1989} is a grid-based representation of the vehicle's environment. Each cell in this grid contains occupancy information. First approaches used a binary Bayes filter to generate an OGM from distance measurements. Recent approaches are often based on evidence theory, which is described in Section~\ref{evidence_theory}. This allows to distinguish unobservable cells from cells with conflicting evidence. Hand-crafted geometric approaches have often been used in the context of OGMs. They assume a ground model to separate drivable from occupied area but have problems in non-flat and dynamic environments. Recent approaches use deep learning models to predict OGMs from distance measurements. Wirges et al. \cite{Wirges.2018} use sequential sparse measurement OGMs to create a dense label OGM to be predicted from one measurement OGM. In \cite{Dequaire.2018}, a model is trained to predict future OGMs using lidar measurements as input data. The approach is still based on a naive model that solely treats all reflection points in a specific height as obstacles. Bauer et al. \cite{Bauer.2019b} transform radar data into an OGM with two channels containing static and dynamic detections. \cite{Filatov.2020} and \cite{Lee.03.08.2020} both use sequential lidar point clouds to create label OGMs including velocity information from tracked or annotated objects. The authors of \cite{Schreiber.2021b} create label OGMs with occupancies estimated by a particle-based approach and velocities from bounding box annotations and train a recurrent neural network to predict dynamic OGMs from measurement OGMs that were created using a geometric ISM.

\subsection{Semantic Grid Maps}

SGMs assign semantic information on the obstacle class to each cell instead of occupancy information. They usually do not quantify the uncertainty of the classification. Wu et al. \cite{Wu.2020} presented a model that predicts SGMs with information on occupancy, object class and motion information in one shot. The input data consists of sequential lidar point clouds represented as binary matrices stacked along a third dimension. Annotations from the nuScenes dataset \cite{Caesar.2020} are used to train the model. The authors of \cite{Bieder.2020}, \cite{Shepel.2021} and \cite{Fei.2021b} combine multiple semantically segmented lidar point clouds from manually annotated datasets to create label SGMs that are to be predicted by the trained model from sparse lidar measurements.

\subsection{Evidence Theory and Subjective Logic} \label{evidence_theory}

In contrast to Bayesian probability theory, in the framework of \textbf{Evidence Theory}~\cite{Shafer.1976} it is possible to consider epistemic uncertainty, which has already been used with OGMs before~\cite{Nuss.2018, Wirges.2018}. Each cell in the OGM contains belief masses $m$ for the possible cell states. In this work, these are the mutually exclusive cell states \textit{free} ($F$), \textit{statically occupied} ($O_s$) and \textit{dynamically occupied} ($O_d$) forming the frame of discernment $\Theta$.
\begin{equation}
\Theta = \{ F,O_s,O_d \}
\end{equation}
During evaluation, we will consider the reduced power set 
\begin{equation}
\mathcal{E} = \{ \{ F \}, \{ O_s \}, \{ O_d \}, \{ O_s, O_d \}, \Theta \}    
\end{equation}
which extends the FOD with the two additional states \textit{statically or dynamically occupied} and \textit{unknown}. The belief mass distribution $m(A)$ can also be interpreted as a subjective opinion in the framework of \textbf{Subjective Logic}~\cite{Jsang.2016}. Motivated by the work of Sensoy et al.~\cite{Sensoy.2018} showing that these parameters can be predicted by a neural network to quantify uncertainty in classification tasks, we created an deep neural network to predict evidential OGMs and use their proposed loss function in our previous~\cite{vanKempen.2021b} and in this work. \textbf{Dempster's Rule of Combination}~\cite{Shafer.2016} can be used to combine evidence from different sources, e.g. evidence for the occupancy state of a grid cell. It allows to compute the joint mass $m_{12}(X)$ for a hypothesis $X$, which is an element in the power sets of the combinable FODs $2^{{\Theta}_1}$ and $2^{{\Theta}_2}$ of the different sources.

\subsection{Synthetic Training Data}

The performance of deep learning models usually drops when presented with test data from a different domain. For perception tasks, this comprises different countries or weather conditions as well as different sensor models or mountings. A domain shift is also caused by the so-called reality gap when a model that has only seen synthetic data during training is presented with test data from the real world. Reiher et al.~\cite{Reiher.2020} presented a methodology for the transformation of camera images into an SGM using synthetic training data. They use segmented camera images predicted by a model trained on real-world data as intermediate representation and thus successfully bridge the reality gap. Zeng et al.~\cite{.2021f} propose a network architecture that improves generalization of lidar object detection models to other domains. They observed that BEV features are more transferable, which is supported by our previous results~\cite{vanKempen.2021b}, where we have shown that it is possible to train a model for evidential occupancy grid mapping from lidar measurements using synthetic data only and obtain qualitatively promising results on real-world test data, as well. In this work, we will present an extension of this algorithm that separates static from dynamic obstacles and evaluate the performance quantitatively on the nuScenes dataset~\cite{Caesar.2020}.

On the other hand, sensor simulation is also improving. While Virtual Test Drive (VTD)~\cite{VirtualTestDrive.2022} uses a physics-based sensor model, the authors of~\cite{Manivasagam.2020} propose a deep neural network that predicts realistic from synthetic lidar points clouds.

\section{DATA-DRIVEN OCCUPANCY GRID MAPPING}

We treat occupancy grid mapping as a supervised learning problem and present our neural network architecture as well as two methods for generating training data. One uses synthetic data generated using simulation and the other one uses labels generated from a manually annotated dataset.

\subsection{Network Architecture}

Our network architecture is based on PointPillars~\cite{Lang.2019b}, which performs well for object detection in lidar point clouds. We use the same feature encoding layer and an adapted version of the CNN backbone. The prediction heads are replaced by our Evidential Prediction Head~\cite{vanKempen.2021b}, which is extended to three layers in this work to comprise evidence for the possible cell states $A \in \Theta$. The model is trained with the Expected Mean Squared Error loss function proposed by Sensoy et al.~\cite{Sensoy.2018}. It is possible to convert evidence $e_{A} \geq 1$ to a subjective opinion $(\boldsymbol{b}, u)$. Here, $K = \left| \Theta \right|$ is the number of classes and $S = \sum_{A \in \Theta} \alpha_{A}$ is the Dirichlet strength.
\begin{eqnarray}
    \alpha_{A} & = & e_{A} + 1 \;, \qquad A \in \Theta \label{eq:evidence_to_alpha} \\
    b_A & = & \frac{e_A}{S} \\
    u & = & \frac{K}{S}
\end{eqnarray}
This belief masses $b_A \in [0,1]$ are used in the evidential OGM predicted by our network.

\subsection{Synthetic Training Data from Simulation}

\begin{figure}
\centering
\subfloat[Synthetic]{
	   \includegraphics[width=0.45\columnwidth]{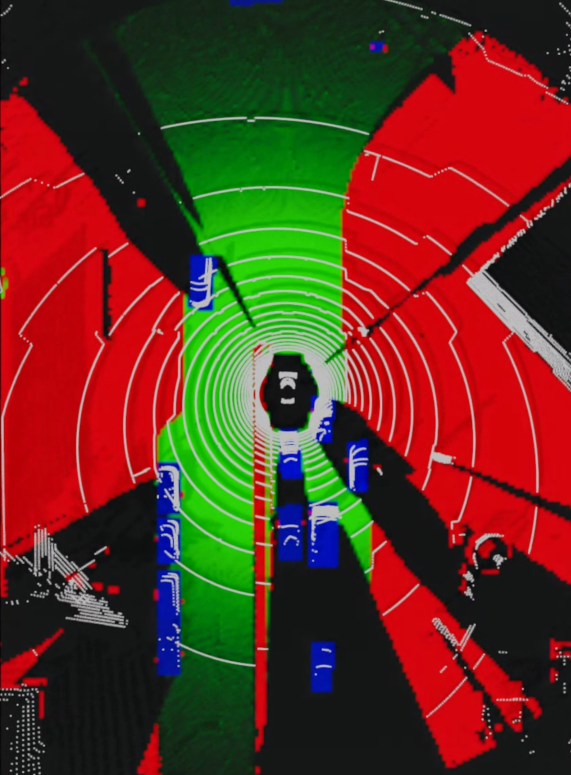}\label{fig:training-sim}
	}
\subfloat[nuScenes]{
	\includegraphics[width=0.45\columnwidth]{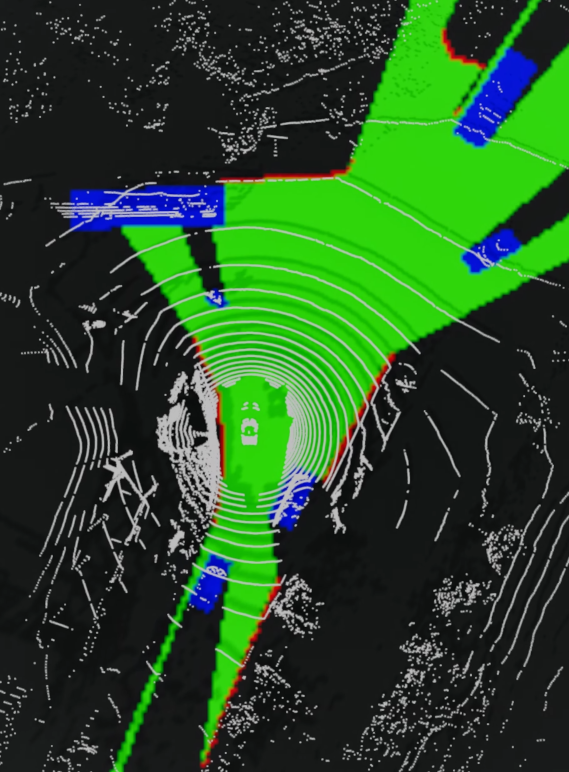}\label{fig:training-nuscenes}
	}
\caption{Training samples are generated using simulation and from the nuScenes dataset~\cite{Caesar.2020}. One sample consists of a lidar point cloud (grey) and an OGM, where green indicates belief mass for \textit{free}, red for \textit{statically occupied} and blue for \textit{dynamically occupied} cells.}
\label{fig:training-samples}
\end{figure}

As already in~\cite{vanKempen.2021b}, we use VTD~\cite{VirtualTestDrive.2022} to create simulation scenarios in an urban 3D environment including static obstacles (e.g. traffic lights, benches, poles) as well as dynamic objects (e.g. pedestrians, bicycles, vehicles) from a large catalogue of vehicle types. The simulated ego vehicle is of a similar size and type as the one used to record the test dataset and uses a vehicle dynamics model to achieve e.g. realistic roll and pitch angles when accelerating or turning. It is equipped with a virtual lidar sensor with a pose similar to the real vehicle. Laser beams are simulated using ray casting in 900 horizontal directions on 32 vertical layers similar to the real sensor. A second virtual lidar sensor with the same pose and field of view but using 3000 instead of 32 vertical layers creates denser lidar point clouds. These synthetic point clouds contain information about the material, which caused the reflection, which is interpreted as evidence for the occupancy state of the grid cell containing the reflection point and its surrounding cells in the label OGMs. In contrast to our previous approach~\cite{vanKempen.2021b}, only reflections on legally drivable material (i.e. asphalt, road marks) contributes a belief mass of $b(F)=0.1$ and all reflections on non-drivable material (e.g. sidewalks, buildings) contribute $m(O_s)=0.1$. By combining all belief masses assigned to one cell using \textit{Dempster's Rule of Combination}, the uncertainty in the created grid map increases with higher distance from the sensor due to a lower density of the lidar rays affecting one cell. In addition to this, a belief mass $m(O_d) = \frac{1}{|C|} \sum_c{m_c(O_s)}$ is assigned to all cells covered by dynamic objects, where $C = \{c_i\}$ is the set of cells occupied by the object. However, information that cannot be inferred from the input data should not be contained in the labels. Thus, only objects that were hit by a minimum of 20 simulated laser beams are processed. One such training sample is shown in Figure \ref{fig:training-sim}.

\subsection{Training Data from Annotations}\label{sec:training-nuscenes}

Our second method uses the manually annotated nuScenes dataset~\cite{Caesar.2020} to create training samples consisting of lidar point clouds and OGMs. While raw lidar measurements can be directly obtained from the dataset, OGM labels have to be generated from the annotations available. We create OGMs of the same size and resolution as in the previous method and assign a belief mass $m(O_d)=1.0$ to all cells covered by objects that are represented by a minimum of 20 points in the lidar point cloud. Additionally, nuScene's map expansion is used to assign a belief mass $m(F)=1.0$ to all cells mapped as drivable surface and not occupied by objects. All other cells are assigned a belief mass $m(O_s)=1.0$. As also in the previous method, we want to respect observability and ensure that labels contain only information that can be inferred from the input data. Using two-dimensional ray casting from the sensor to the grid borders, all cells behind static or dynamic obstacles are assigned a maximum uncertainty $m(\Theta)=1$, with the exception of all cells covered by dynamic objects as it is assumed that their size can be inferred from the input data if they are represented by at least 20 reflection points in the measurement. An example of such a label is shown in Figure \ref{fig:training-nuscenes}.

\section{EXPERIMENTAL SETUP AND RESEARCH QUESTIONS}

Using the methods described above, $10.000$ synthetic training samples as well as $30.000$ samples using nuScenes' training split were generated. We used the same neural network architecture as described in~\cite{vanKempen.2021b} to predict OGMs of $81.92\,m$ length, $56.32\,m$ width and a cell size of $32\,cm \times 32\,cm$. A minimum loss was achieved after approx. $50$ epochs of training for both models. Their performance is evaluated on $6000$ samples from nuScene's validation split using the metrics described next. The ground-truth OGMs for evaluation are created as described in Section \ref{sec:training-nuscenes}. As is apparent in Figure \ref{fig:training-samples}, the synthetic labels contain significantly more static cells than the labels created from nuScenes. To get comparable results, only cells with a known state (i.e. $m(\Theta) < 0.5$) in the ground-truth OGM are evaluated. As there is no uncertainty quantification in these labels, we use a threshold of $m_\vartheta = 0.5$ to classify cells in the predicted OGM into classes \textit{free}, \textit{statically occupied} and \textit{dynamically occupied}. This allows us to compute the \textbf{Precision} $P_A = \frac{TP}{TP + FP}$ and \textbf{Recall} $R_A = \frac{TP}{TP+FN}$ from the numbers of true positive (TP), false positive (FP) and false negative (FN) cell classifications for the relevant states $A \in \mathcal{E}$. By analyzing these metrics we want to answer the following research questions:
\begin{itemize}
    \item Is it possible to train a deep neural network to predict an evidential OGM, which distinguishes between statically and dynamically occupied areas, when presented with a single lidar measurement?
    \item If such a model was trained with synthetic data only, is it possible to perform a domain shift and achieve a sufficient performance on real-world data, as well?
    \item If such a model was trained with annotated data, how well does it perform when presented with unseen data from the same dataset?
    \item Which of both approaches generalizes better to new sensor configurations, i.e. which model performs better when presented with input data from another sensor on another vehicle?
\end{itemize}

\section{RESULTS AND DISCUSSION}

First, we analyze the performance of both models quantitatively on validation data from the nuScenes~\cite{Caesar.2020} dataset. Then we will perform a domain shift and use both approaches with lidar measurements recorded with our research vehicle to evaluate the generalization capability of both approaches qualitatively. \textit{Model A} is trained using synthetic data and \textit{model B} is trained using labels generated from annotations.

\begin{table}
    \centering
    \begin{tabular}{|c | c c | c c | c c | c c|}
     \hline
      & $P_F$ & $R_F$ & $P_{O_s}$ & $R_{O_s}$ & $P_{O_d}$ & $R_{O_d}$ & $P_{O_{sd}}$ & $R_{O_{sd}}$ \\[1ex]
     \hline\hline
     A & 0.98 & 0.34 & 0.06 & 0.44 & 0.55 & 0.64 & 0.23 & 0.73 \\ 
     \hline
     B & 0.99 & 0.71 & 0.33 & 0.45 & 0.85 & 0.82 & 0.64 & 0.72 \\ 
     \hline
    \end{tabular}\caption{Quantitative evaluation of both approaches on unseen real-world data. \textit{Model A} uses synthetic training data and \textit{model B} uses labels created from manual annotations.}\label{tab:results}
\end{table}

\subsection{Evaluation on nuScenes}\label{sec:eval-nuscenes}

Figure~\ref{fig:predictions} shows predictions of both trained models on lidar measurements from unseen data of nuScene's validation split. Both models successfully predict evidential OGMs and take observability into account. Both approaches have also learned to estimate \textit{statically occupied} and \textit{dynamically occupied} areas where also the dimensions of dynamic objects are being estimated. The most noticeable difference is the number of \textit{statically occupied} cells, which results from the different methods for creating the labels. As the labels from synthetic data are created using a simulated lidar sensor, the field of view can be estimated better than with the ray casting approach used to create labels from annotations. This is also reflected in the quantitative evaluation and explains the low precision for statically occupied cells in Table~\ref{tab:results}. The table summarizes the precision and recall for the cell states \textit{free} ($F$), \textit{statically occupied} ($O_s$), \textit{dynamically occupied} ($O_d$) and for the combined state \textit{statically or dynamically occupied} ($O_{sd}$) that were achieved with both approaches on the validation data. A high precision of $0.98$ for \textit{model A} and $0.99$ for \textit{model B} is achieved for free space estimation, while the recall is considerably lower. A qualitative analysis shows that free space is missed especially in higher distance from the sensor where lidar measurements are only sparse. This makes it hard to estimate the cell states from a single lidar measurement and could be solved by tracking the OGM over time. The higher recall of \textit{model B} can be explained by the higher diversity of road conditions in real-world data. E.g. elevation is not represented yet in the simulated scenarios so that free space cannot be estimated correctly at inclinations. While both models have a similar recall when combining statically and dynamically occupied cells, \textit{model B} can better distinguish dynamic from static obstacles. This can be explained with the higher diversity of dynamic and static objects in the real world. However, the evaluation on an annotated dataset allows to detect missed object types so that they can also be modeled in simulation.

\subsection{Evaluation of Generalization After Domain Shift}

\begin{figure}[h]
\centering
\subfloat[Trained on Synthetic Data]{
	   \includegraphics[width=0.45\columnwidth]{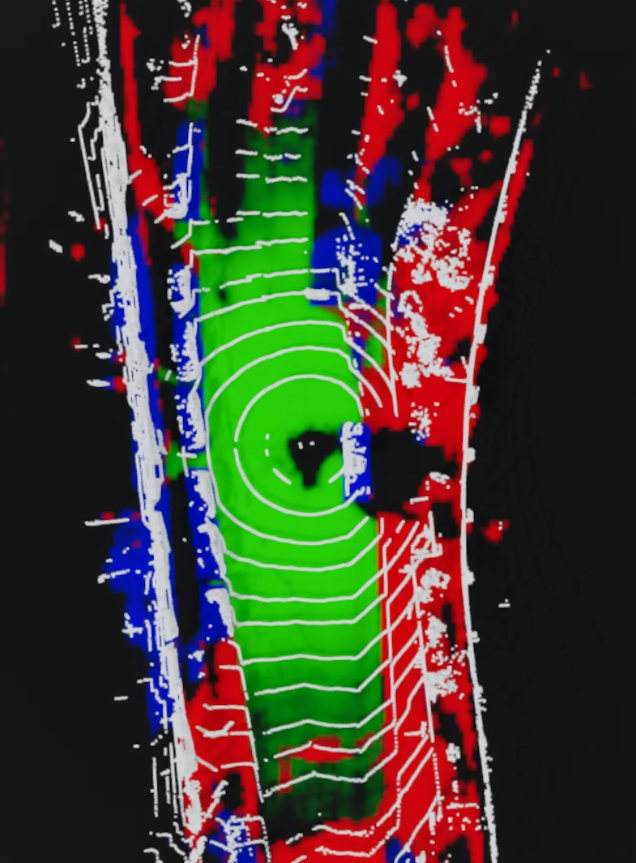}\label{fig:prediction-vlp32-syn}
	}
\subfloat[Trained on nuScenes]{
	\includegraphics[width=0.45\columnwidth]{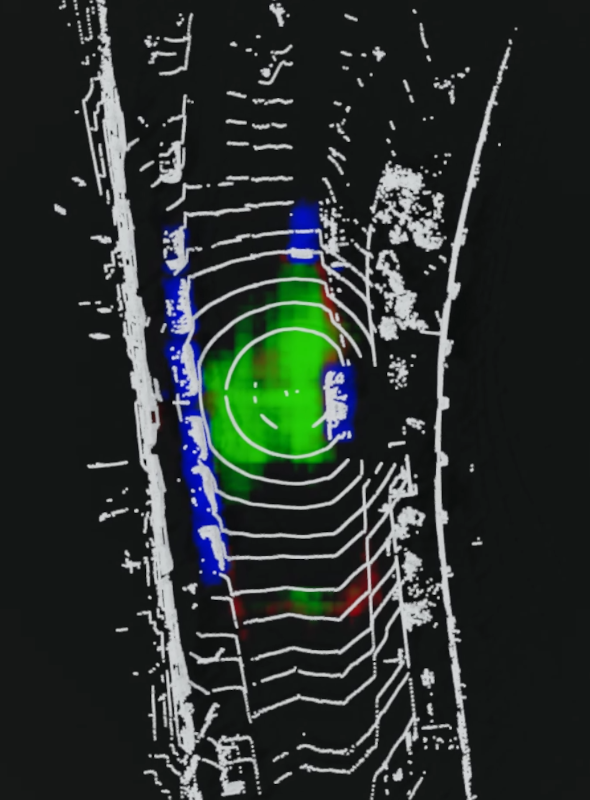}\label{fig:prediction-vlp32-nuscenes}
	}
\caption{Both approaches are tested with data from a \textit{Velodyne VLP32C} lidar sensor on one of our test vehicles. The model predicting the left OGM was trained using new synthetic data while the right OGM is predicted by a model trained with annotations from the nuScenes dataset.}
\label{fig:predictions-vlp32}
\end{figure}

We also want to analyze the ability of both presented methods to generalize to a new domain, in particular when used with a new sensor configuration. We tested both approaches our research vehicle having a \textit{Velodyne VLP32C} mounted on its roof. For the approach relying on synthetic training data, it is possible to generate new training data using a simulated vehicle and sensor setup similar to our research vehicle with low effort. This is not the case for a manual annotated dataset so that the performance of a model trained on this data is compared to the model trained on annotations from nuScenes. Figure~\ref{fig:predictions-vlp32} shows predictions of both models. It is apparent that the distribution of reflection points differs from the sensor used in the nuScenes dataset (cf. Figure~\ref{fig:prediction-nuscenes}). The model trained on data from this dataset shows a substantially lower performance when confronted with a domain shift, while the model trained on new synthetic data shows similar results in both domains (cf. Figure~\ref{fig:prediction-sim}).

\section{CONCLUSION}

We presented two data-driven methods for occupancy grid mapping from lidar measurements. One approach uses synthetic training data while the other approach uses labels generated from manual annotations of the nuScenes~\cite{Caesar.2020} dataset. An evaluation on real-world data shows that both approaches are able to predict evidential OGMs that distinguish between static and potentially dynamic obstacles and take observability into account. A quantitative analysis shows that the model trained on synthetic data achieves lower performance on separating static from dynamic obstacles while free space is estimated with high precision with both approaches. However, the model trained on the annotated dataset performs significantly worse when presented with lidar measurements from another sensor model on another vehicle while the approach using synthetic data can be easily transferred. By an evaluation on annotated real-world data it becomes possible to identify the blind spots of simulation, e.g. falsely classified static and dynamic obstacles or road layouts. This information can be used to improve the simulation and further close the reality gap so that synthetic training data can be generated for arbitrary sensor configurations.


\bibliographystyle{IEEEtran}
\bibliography{literature}

\begin{thebibliography}{10}
\providecommand{\url}[1]{#1}
\csname url@samestyle\endcsname
\providecommand{\newblock}{\relax}
\providecommand{\bibinfo}[2]{#2}
\providecommand{\BIBentrySTDinterwordspacing}{\spaceskip=0pt\relax}
\providecommand{\BIBentryALTinterwordstretchfactor}{4}
\providecommand{\BIBentryALTinterwordspacing}{\spaceskip=\fontdimen2\font plus
\BIBentryALTinterwordstretchfactor\fontdimen3\font minus
  \fontdimen4\font\relax}
\providecommand{\BIBforeignlanguage}[2]{{%
\expandafter\ifx\csname l@#1\endcsname\relax
\typeout{** WARNING: IEEEtran.bst: No hyphenation pattern has been}%
\typeout{** loaded for the language `#1'. Using the pattern for}%
\typeout{** the default language instead.}%
\else
\language=\csname l@#1\endcsname
\fi
#2}}
\providecommand{\BIBdecl}{\relax}
\BIBdecl

\bibitem{Caesar.2020}
H.~Caesar, V.~Bankiti \emph{et~al.}, ``nuscenes: A multimodal dataset for
  autonomous driving,'' in \emph{2020 IEEE/CVF Conference on Computer Vision
  and Pattern Recognition}, 2020, pp. 11\,618--11\,628.

\bibitem{Buchholz.2020}
M.~Buchholz, F.~Gies \emph{et~al.}, ``Automation of the unicaragil vehicles,''
  in \emph{29th Aachen Colloquium}, 2020, pp. 1531--1560.

\bibitem{Woopen.2018}
T.~Woopen, B.~Lampe \emph{et~al.}, ``Unicaragil - disruptive modular
  architectures for agile, automated vehicle concepts,'' in \emph{27th Aachen
  Colloquium}, 2018, pp. 663--694.

\bibitem{Bieder.2020}
F.~Bieder, S.~Wirges \emph{et~al.}, ``Exploiting multi-layer grid maps for
  surround-view semantic segmentation of sparse lidar data,'' in \emph{2020
  IEEE Intelligent Vehicles Symposium (IV)}.\hskip 1em plus 0.5em minus
  0.4em\relax IEEE, 2020.

\bibitem{Richter.2020}
S.~Richter, J.~Beck \emph{et~al.}, ``Semantic evidential grid mapping based on
  stereo vision,'' in \emph{2020 IEEE International Conference on Multisensor
  Fusion and Integration for Intelligent Systems (MFI)}.\hskip 1em plus 0.5em
  minus 0.4em\relax IEEE, 2020.

\bibitem{vanKempen.2021b}
R.~{van Kempen}, B.~Lampe \emph{et~al.}, ``A simulation-based end-to-end
  learning framework for evidential occupancy grid mapping,'' in \emph{2021
  IEEE Intelligent Vehicles Symposium (IV)}, 2021, pp. 934--939.

\bibitem{Lampe.2020}
B.~Lampe, R.~{van Kempen} \emph{et~al.}, ``Reducing uncertainty by fusing
  dynamic occupancy grid maps in a cloud-based collective environment model,''
  in \emph{2020 IEEE Intelligent Vehicles Symposium (IV)}, 2020, pp. 837--843.

\bibitem{Thrun.2005}
S.~Thrun, W.~Burgard, and D.~Fox, \emph{Probabilistic robotics}, ser.
  Intelligent robotics and autonomous agents.\hskip 1em plus 0.5em minus
  0.4em\relax Cambridge, Mass.: {MIT Press}, 2005.

\bibitem{Nuss.2018}
D.~Nuss, S.~Reuter \emph{et~al.}, ``A random finite set approach for dynamic
  occupancy grid maps with real-time application,'' \emph{The International
  Journal of Robotics Research}, vol.~37, no.~8, pp. 841--866, 2018.

\bibitem{Bauer.2019b}
D.~Bauer, L.~Kuhnert, and L.~Eckstein, ``Deep, spatially coherent inverse
  sensor models with uncertainty incorporation using the evidential
  framework,'' in \emph{2019 IEEE Intelligent Vehicles Symposium (IV)}, 2019,
  pp. 2490--2495.

\bibitem{Lee.03.08.2020}
K.-H. Lee, M.~Kliemann \emph{et~al.}, ``Pillarflow: End-to-end birds-eye-view
  flow estimation for autonomous driving,'' 03.08.2020.

\bibitem{Fei.2021b}
J.~Fei, K.~Peng \emph{et~al.}, ``Pillarsegnet: Pillar-based semantic grid map
  estimation using sparse lidar data,'' in \emph{2021 IEEE Intelligent Vehicles
  Symposium (IV)}, 2021, pp. 838--844.

\bibitem{Schreiber.2021b}
M.~Schreiber, V.~Belagiannis \emph{et~al.}, ``Dynamic occupancy grid mapping
  with recurrent neural networks,'' in \emph{2021 IEEE International Conference
  on Robotics and Automation (ICRA)}, 2021, pp. 6717--6724.

\bibitem{Elfes.1989}
A.~Elfes, ``Using occupancy grids for mobile robot perception and navigation,''
  \emph{Computer}, vol.~22, no.~6, pp. 46--57, 1989.

\bibitem{Wirges.2018}
S.~Wirges, C.~Stiller, and F.~Hartenbach, ``Evidential occupancy grid map
  augmentation using deep learning,'' in \emph{2018 IEEE Intelligent Vehicles
  Symposium (IV 2018)}.\hskip 1em plus 0.5em minus 0.4em\relax IEEE, 2018, pp.
  668--673.

\bibitem{Dequaire.2018}
J.~Dequaire, P.~Ondr{\'u}{\v{s}}ka \emph{et~al.}, ``Deep tracking in the wild:
  End-to-end tracking using recurrent neural networks,'' \emph{The
  International Journal of Robotics Research}, vol.~37, no. 4-5, pp. 492--512,
  2018.

\bibitem{Filatov.2020}
A.~Filatov, A.~Rykov, and V.~Murashkin, ``Any motion detector: Learning
  class-agnostic scene dynamics from a sequence of lidar point clouds,'' in
  \emph{2020 IEEE International Conference on Robotics and Automation
  (ICRA)}.\hskip 1em plus 0.5em minus 0.4em\relax IEEE, 2020, pp. 9498--9504.

\bibitem{Wu.2020}
P.~Wu, S.~Chen, and D.~N. Metaxas, ``Motionnet: Joint perception and motion
  prediction for autonomous driving based on bird's eye view maps,'' in
  \emph{2020 IEEE/CVF Conference on Computer Vision and Pattern Recognition},
  2020, pp. 11\,382--11\,392.

\bibitem{Shepel.2021}
I.~Shepel, V.~Adeshkin \emph{et~al.}, ``Occupancy grid generation with dynamic
  obstacle segmentation in stereo images,'' \emph{IEEE Transactions on
  Intelligent Transportation Systems}, pp. 1--11, 2021.

\bibitem{Shafer.1976}
G.~Shafer, \emph{A mathematical theory of evidence}, ser. Limited paperback
  editions.\hskip 1em plus 0.5em minus 0.4em\relax Princeton, NJ: {Princeton
  Univ. Press}, 1976, vol.~42.

\bibitem{Jsang.2016}
A.~J{\o}sang, \emph{Subjective Logic}.\hskip 1em plus 0.5em minus 0.4em\relax
  Cham: {Springer International Publishing}, 2016.

\bibitem{Sensoy.2018}
M.~Sensoy, L.~Kaplan, and M.~Kandemir, ``Evidential deep learning to quantify
  classification uncertainty,'' in \emph{Advances in Neural Information
  Processing Systems 31 (NeurIPS 2018)}, 2018, pp. 3179--3189.

\bibitem{Shafer.2016}
G.~Shafer, ``Dempster's rule of combination,'' \emph{International Journal of
  Approximate Reasoning}, vol.~79, pp. 26--40, 2016.

\bibitem{Reiher.2020}
L.~Reiher, B.~Lampe, and L.~Eckstein, ``A sim2real deep learning approach for
  the transformation of images from multiple vehicle-mounted cameras to a
  semantically segmented image in bird's eye view,'' in \emph{2020 IEEE 23rd
  International Conference on Intelligent Transportation Systems (ITSC)}, 2020,
  pp. 1--7.

\bibitem{.2021f}
{Zeng Yihan, Chunwei Wang, Yunbo Wang, Hang Xu, Chaoqiang Ye, Zhen Yang, Chao
  Ma}, ``Learning transferable features for point cloud detection via 3d
  contrastive co-training,'' in \emph{Advances in Neural Information Processing
  Systems}, 2021, vol.~34.

\bibitem{VirtualTestDrive.2022}
\BIBentryALTinterwordspacing
{VIRES Simulationstechnologie GmbH}, ``Virtual test drive,'' 2022. [Online].
  Available: \url{https://vires.mscsoftware.com/}
\BIBentrySTDinterwordspacing

\bibitem{Manivasagam.2020}
S.~Manivasagam, S.~Wang \emph{et~al.}, ``Lidarsim: Realistic lidar simulation
  by leveraging the real world,'' in \emph{2020 IEEE/CVF Conference on Computer
  Vision and Pattern Recognition (CVPR)}, 2020, pp. 11\,164--11\,173.

\bibitem{Lang.2019b}
A.~H. Lang, S.~Vora \emph{et~al.}, ``Pointpillars: Fast encoders for object
  detection from point clouds,'' in \emph{2019 IEEE/CVF Conference on Computer
  Vision and Pattern Recognition (CVPR)}, 2019, pp. 12\,689--12\,697.

\end{thebibliography}

\vspace{12pt}

\end{document}